\documentclass[10pt,twocolumn,letterpaper]{article}

\usepackage{cvpr}              
\usepackage[accsupp]{axessibility}
\usepackage{graphicx} 
\usepackage{float} 
\usepackage{cuted}
\usepackage{caption}
\usepackage{multirow}
\usepackage{pifont}

\newcommand{\xmark}{\ding{55}} 
\definecolor{cvprblue}{rgb}{0.21,0.49,0.74}
\usepackage[pagebackref,breaklinks,colorlinks,allcolors=cvprblue]{hyperref}

\def\paperID{93} 
\def\confName{CVPR}
\def\confYear{2026}

\title{Degradation-Aware and Structure-Preserving Diffusion for Real-World Image Super-Resolution}

\author{Yang Ji\\
KUNBYTE\\
Hangzhou, Zhejiang Province, China\\
{\tt\small jiyang\_qut@163.com}
\and
Zonghao Chen\\
KUNBYTE\\
Hangzhou, Zhejiang Province, China\\
{\tt\small chenzonghao@kunbyte.com}
\and
Zhihao Xue\\
KUNBYTE\\
Hangzhou, Zhejiang Province, China\\
{\tt\small xuezhihao@kunbyte.com}
\and
Junqin Hu\\
GOLDMYE\\
Xuhui District, Shanghai, China\\
{\tt\small hujunqin@goldmye.com}
}

\begin{document}

\maketitle

\begin{strip}
    \centering
    \includegraphics[width=\textwidth]{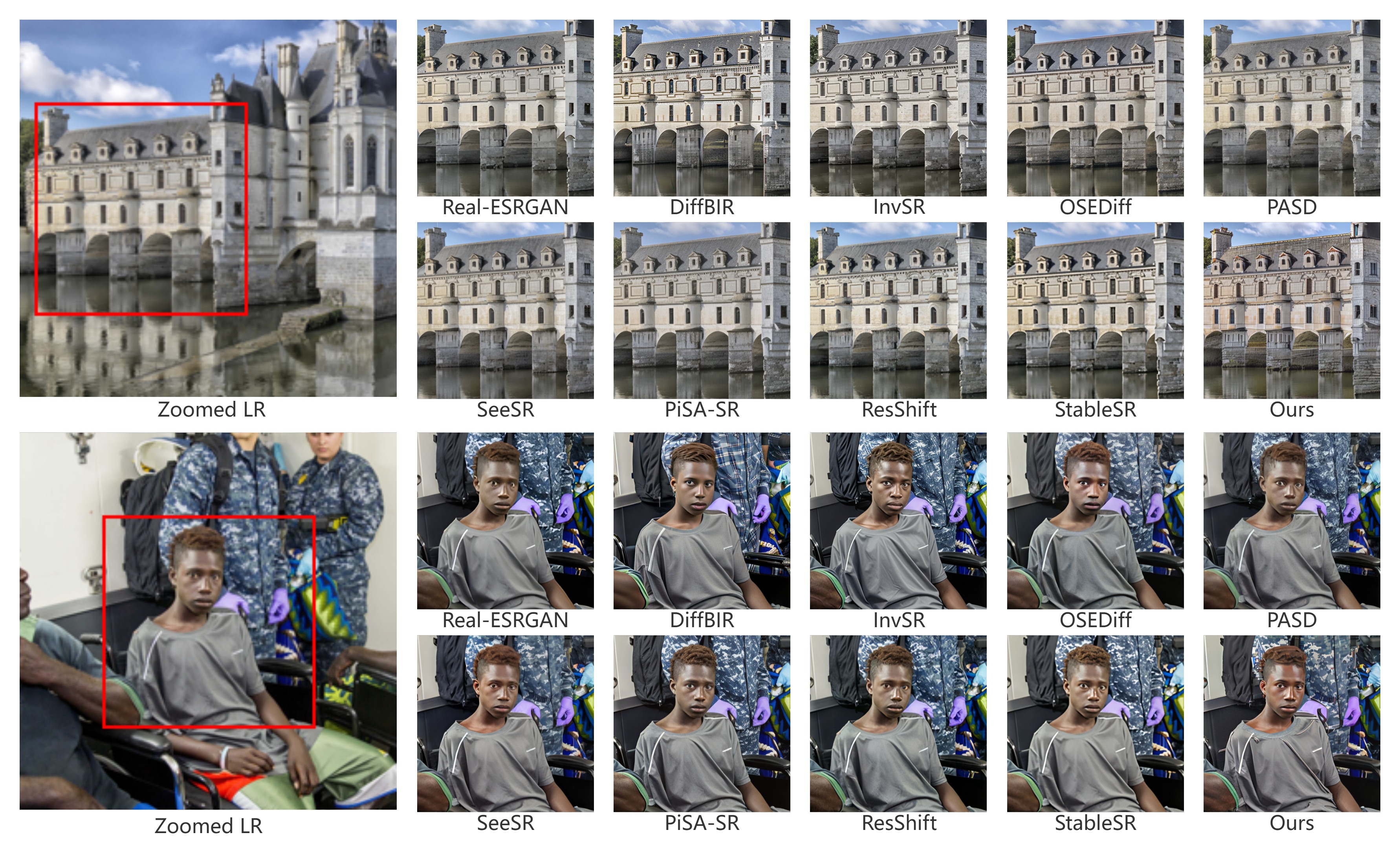}
    \captionof{figure}{Visual results of different methods on two typical real-world examples from DIV2K dataset.} 
    \label{fig:qualitative_1}
    \vspace{5pt} 
\end{strip}

\begin{abstract}
Real-world image super-resolution is particularly challenging for diffusion models because real degradations are complex, heterogeneous, and rarely modeled explicitly. We propose a degradation-aware and structure-preserving diffusion framework for real-world SR. Specifically, we introduce Degradation-aware Token Injection, which encodes lightweight degradation statistics from low-resolution inputs and fuses them with semantic conditioning features, enabling explicit degradation-aware restoration. We further propose Spatially Asymmetric Noise Injection, which modulates diffusion noise with local edge strength to better preserve structural regions during training. Both modules are lightweight add-ons to the adopted diffusion SR framework, requiring only minor modifications to the conditioning pipeline. Experiments on DIV2K and RealSR show that our method delivers competitive no-reference perceptual quality and visually more realistic restoration results than recent baselines, while maintaining a favorable perception--distortion trade-off. Ablations confirm the effectiveness of each module and their complementary gains when combined. The code and model are publicly available at \href{https://github.com/jiyang0315/DASP-SR.git}{https://github.com/jiyang0315/DASP-SR}.
\end{abstract}
\section{Introduction}
\label{sec:intro}

Single image super-resolution (SISR) is a fundamental challenge in computer vision, aiming to reconstruct a high-resolution (HR) image from its low-resolution (LR) counterpart. While traditional learning-based methods driven by pixel-wise regression achieve high PSNR~\cite{SRCNN,VDSR,EDSR}, they often suffer from regression-to-the-mean, resulting in over-smoothed textures that lack perceptual realism~\cite{SRGAN,PerceptionDistortion}. Recently, diffusion probabilistic models have emerged as a powerful paradigm for generative modeling~\cite{DDPM,IDDPM}, demonstrating an exceptional ability to synthesize high-frequency details for image super-resolution~\cite{SR3}. However, transitioning diffusion models to real-world SR remains non-trivial. Unlike classical benchmarks that assume a fixed bicubic downsampling, real-world images are plagued by complex, unknown, and spatially varying degradations, including blur, noise, and compression artifacts, which pose significant hurdles for robust restoration~\cite{BSRGAN,RealESRGAN,PDARWSR}.

Existing diffusion-based SR methods typically condition the denoising process on the LR image via simple concatenation or latent embeddings~\cite{SR3,SRDiff,StableSR}. Despite their improved visual quality, these approaches largely treat the LR input as a monolithic conditioning signal, failing to explicitly decouple the underlying degradation patterns from the image content. This implicit treatment often leads to restoration ambiguity; without explicit knowledge of the degradation type and severity, the model may struggle to distinguish between genuine textures and artifacts, resulting in either insufficient restoration or unnatural hallucinations. Recent methods such as SeeSR and DiffBIR partially alleviate this issue by introducing degradation-aware conditioning and restoration guidance~\cite{SeeSR,DiffBIR}. Furthermore, standard diffusion training applies spatially uniform Gaussian noise to the latent space~\cite{DDPM,LDM}. This overlooks the inherent spatial heteroscedasticity of image information: high-frequency structural regions (e.g., edges) are more sensitive to noise perturbations than flat areas. Applying uniform noise tends to prematurely eradicate critical structural cues, forcing the model to "re-invent" geometry from scratch rather than refining existing structural remnants.

To address these limitations, we propose a degradation-aware and structure-preserving diffusion framework for real-world SISR. Rather than introducing a fundamentally new diffusion formulation, our approach focuses on practical and targeted improvements to conditioning and training noise design within the adopted diffusion SR framework. Specifically, we introduce two lightweight modules that provide explicit restoration guidance. First, we develop a Degradation-aware Token Injection (DTI) module. DTI extracts a compact set of degradation statistics, encompassing blur kernels, noise levels, compression severity, and contrast, and projects them into a specialized conditioning token. By fusing this token with semantic embeddings, we provide the denoising network with explicit degradation cues, enabling more targeted and robust reconstruction across diverse degradation scenarios. Second, we introduce a Spatially Asymmetric Noise Injection (SANI) strategy. Unlike standard uniform diffusion, SANI modulates the training noise magnitude based on local edge strength. By assigning lower noise levels to structural regions, SANI preserves the geometric skeleton of the image during the diffusion process, effectively anchoring the generative process to the original structural topology.

We evaluate our method on the DIV2K and RealSR benchmarks~\cite{DIV2K,RealSR} using a comprehensive suite of full-reference (PSNR, SSIM~\cite{SSIM}, LPIPS) and no-reference perceptual metrics (NIQE~\cite{NIQE}, NRQM, PI~\cite{PI}, CLIP-IQA~\cite{CLIPIQA}, MUSIQ~\cite{MUSIQ}). Experimental results demonstrate that our method achieves a favorable trade-off between perceptual quality and fidelity, with competitive or superior performance on no-reference perceptual metrics compared with recent baselines. Extensive ablation studies further verify that DTI and SANI provide complementary benefits within the adopted diffusion SR pipeline.In addition, this work is part of the NTIRE 2026 Image Super-Resolution ($\times4$) challenge, whose benchmark results and method overview are summarized in the corresponding challenge report~\cite{ntire26srx4}.

Our main contributions are summarized as follows:
\begin{itemize}
\item We propose Degradation-aware Token Injection (DTI), which explicitly encodes degradation characteristics into conditioning tokens, enabling the diffusion model to better handle diverse real-world corruptions through more targeted restoration conditioning.
\item We introduce Spatially Asymmetric Noise Injection (SANI), a structure-aware training strategy that modulates noise levels based on local image geometry to better preserve edges and fine textures.
\item Extensive experiments on DIV2K and RealSR demonstrate that the proposed modules consistently improve perceptual quality and structural integrity on the adopted diffusion SR framework, with only marginal parameter and runtime overhead.
\end{itemize}
\section{Related Work}
\label{sec:related_work}

\noindent\textbf{Real-World Image Super-Resolution.}
Image super-resolution has evolved from pixel-wise regression toward perceptual realism. SRGAN~\cite{SRGAN} and ESRGAN~\cite{ESRGAN} pioneer this direction by combining adversarial and perceptual losses, but their effectiveness is largely limited to synthetic bicubic degradations. To handle complex real-world corruptions, BSRGAN~\cite{BSRGAN} introduces a shuffled degradation pipeline to cover diverse corruption types, and Real-ESRGAN~\cite{RealESRGAN} further advances robustness through high-order degradation modeling. Recognizing that real images often suffer from non-uniform corruptions, recent works explore spatially-variant degradation models that estimate per-pixel degradation patterns~\cite{SpatiallyVariantSR1,SpatiallyVariantSR2}, while CDFormer~\cite{CDFormer} learns content-aware degradation priors via a diffusion-based estimator. Despite these advances, most methods still characterize degradations implicitly through latent embeddings or complex synthesis pipelines, leaving the potential of explicit, interpretable degradation cues largely unexplored.

\noindent\textbf{Diffusion Models for Image Restoration.}
Diffusion probabilistic models have emerged as powerful generative priors for image restoration, owing to their strong ability to synthesize realistic textures and fine structural details~\cite{DDPM,SR3,DiffusionRestoration1,DiffusionRestoration2}. In real-world SR, StableSR~\cite{StableSR} and PASD~\cite{PASD} leverage pre-trained latent diffusion models to inherit rich visual priors, producing perceptually realistic outputs. DiffBIR~\cite{DiffBIR} extends this paradigm to blind image restoration by decoupling degradation removal and detail regeneration into two complementary stages. To address the prohibitive inference cost of standard diffusion sampling, ResShift~\cite{ResShift} reformulates the forward process via residual shifting, achieving competitive quality with substantially fewer steps, and SinSR~\cite{SinSR} further pushes efficiency toward one-step diffusion SR. SeeSR~\cite{SeeSR} introduces degradation-aware semantic prompts to maintain semantic correctness under severe degradations. These results collectively confirm that diffusion models provide a versatile generative foundation for real-world SR, though how to effectively condition the generation process remains a key open problem.

\noindent\textbf{Conditioning and Degradation-Aware Guidance.}
The generative flexibility of diffusion models also introduces the risk of hallucination and semantic drift, making effective conditioning and guidance mechanisms essential. SeeSR~\cite{SeeSR} addresses this by extracting hard and soft semantic prompts that remain reliable under heavy corruption, while DiffBIR~\cite{DiffBIR} introduces region-adaptive restoration guidance to balance realism and fidelity. PASD~\cite{PASD} further complements semantic conditioning with pixel-level structural guidance. BlindDiff~\cite{BlindDiff} takes a more direct approach by integrating degradation kernel estimation into the diffusion sampling loop, enabling kernel-aware generation. More broadly, recent studies~\cite{LearningDiffusionTexturePriors, IterativelyPreconditionedGuidance} demonstrate that carefully designed texture priors and iteratively preconditioned guidance are critical for closing the perception--fidelity gap. However, existing conditioning strategies either entangle degradation and content representations, or rely on heavy auxiliary modules for degradation estimation. In this work, we inject lightweight and interpretable degradation descriptors directly into the diffusion conditioning process, and further incorporate a structure-aware control mechanism to preserve geometric coherence, achieving better degradation adaptability and structural faithfulness with minimal additional overhead.
\section{Method}
\label{sec:method}

\subsection{Overview}
\begin{figure*}[ht]
    \centering
    \includegraphics[width=1\textwidth]{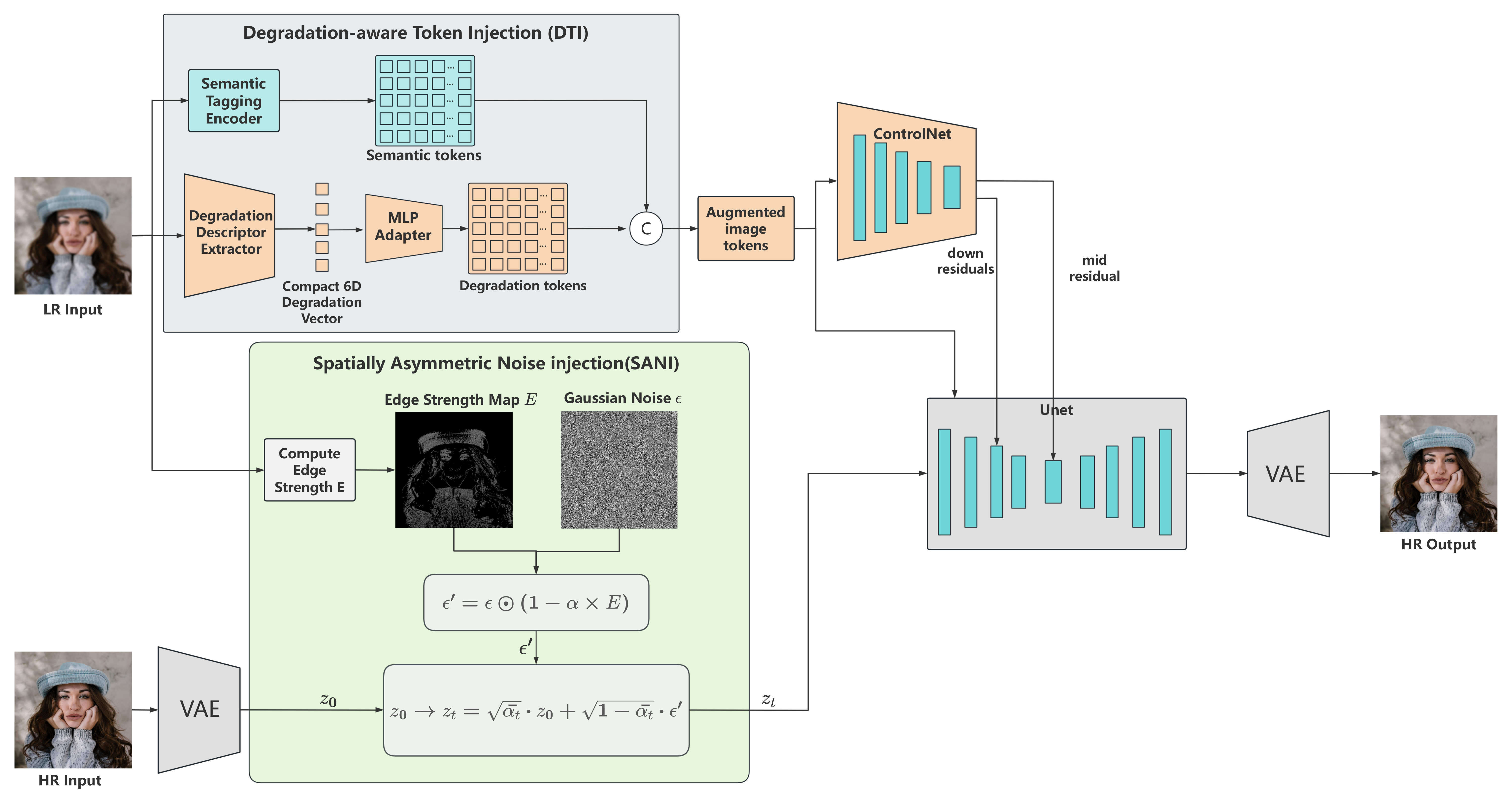}
    \caption{Overview of the proposed framework. The DTI module extracts a six-dimensional degradation descriptor from the LR input and injects it as an additional token into the image cross-attention branch of ControlNet and UNet. The SANI module spatially modulates training noise according to an edge-strength map, reducing perturbation in structure-rich regions.}
    \label{fig:framework}
\end{figure*}

Given a low-resolution image $\mathbf{x}_{\mathrm{lr}} \in \mathbb{R}^{3 \times H/s \times W/s}$ and its high-resolution target $\mathbf{x}_{\mathrm{hr}} \in \mathbb{R}^{3 \times H \times W}$, where $s$ denotes the upsampling factor, our goal is to restore a perceptually faithful SR result under complex real-world degradations. Let $\mathbf{z}_0 = \mathcal{E}(\mathbf{x}_{\mathrm{hr}}) \in \mathbb{R}^{C \times h \times w}$ denote the latent encoding of the HR image, where $\mathcal{E}$ is the VAE encoder and $(h, w)$ is the latent resolution. We introduce two lightweight add-on modules on top of the adopted diffusion SR backbone: (1)~\textbf{Degradation-aware Token Injection (DTI)}, which explicitly encodes degradation statistics from the LR input and injects them into the image cross-attention branch; and (2)~\textbf{Spatially Asymmetric Noise Injection (SANI)}, which modulates diffusion training noise according to local edge strength to better preserve structural regions. DTI exposes explicit degradation cues to the model for more accurate restoration conditioning, while SANI reduces excessive perturbation in structure-rich regions during training. The overall pipeline is illustrated in Fig.~\ref{fig:framework}.

\subsection{Degradation-aware Token Injection}
Existing diffusion-based SR methods condition the denoising process on textual or semantic cues but do not explicitly account for the degradation characteristics of the LR input. To address this, we extract a compact degradation descriptor $\mathbf{d} \in \mathbb{R}^{6}$ from $\mathbf{x}_{\mathrm{lr}}$:
\begin{equation}
\begin{split}
  \mathbf{d} = \log \Bigl( 1 + [ & d_{\mathrm{blur}}, d_{\mathrm{noise}}, d_{\mathrm{jpeg}}, \\
  & d_{\mathrm{edge}}, d_{\mathrm{bright}}, d_{\mathrm{contrast}} ]^{\top} \Bigr) \in \mathbb{R}^{6},
\end{split}
\label{eq:deg_descriptor}
\end{equation}
where all features are computed from the grayscale image $I_g = 0.299R + 0.587G + 0.114B$. The six components capture the primary degradation axes commonly encountered in real-world SR: $d_{\mathrm{blur}} = 1/(\mathrm{Var}(\nabla^2 I_g) + \varepsilon)$ measures blur via the reciprocal of Laplacian variance; $d_{\mathrm{noise}}$ is the mean absolute residual after $3\times3$ average pooling; $d_{\mathrm{jpeg}}$ captures $8\times8$ blocking artifacts from boundary discontinuities; $d_{\mathrm{edge}}$ is the fraction of pixels whose Sobel gradient magnitude exceeds $0.08$; and $d_{\mathrm{bright}}$, $d_{\mathrm{contrast}}$ are the grayscale mean and standard deviation, respectively. The $\log(1+\cdot)$ transform stabilizes the range across diverse degradation levels.

\noindent\textbf{Design rationale of the degradation descriptor.}
The six components are chosen to cover the primary degradation axes that commonly arise in real-world SR and that can be reliably estimated from a single LR image without any learned prior: blur severity (from Laplacian variance), additive noise level (from local residuals), JPEG blocking (from $8\times8$ boundary discontinuities), edge density (from Sobel gradients), and global luminance statistics (brightness and contrast). Together, these statistics span the most frequent degradation types simulated by standard real-world SR pipelines~\cite{RealESRGAN,BSRGAN}. The lightweight, interpretable nature of this descriptor ensures robustness and computational efficiency. We acknowledge that for highly complex or novel mixed degradations beyond these axes, a learned descriptor may offer greater representational capacity; we leave this extension to future work.

We map $\mathbf{d}$ into the image cross-attention space through a lightweight degradation token adapter: a two-layer MLP ($\mathbb{R}^6 \to \mathbb{R}^{128} \to \mathbb{R}^D$) with SiLU activation and dropout ($p{=}0.1$), followed by LayerNorm:
\begin{equation}
  \mathbf{t}_{\mathrm{deg}} = \mathrm{LN}\bigl(\mathrm{MLP}(\mathbf{d})\bigr)
  \in \mathbb{R}^{D},
  \label{eq:deg_token}
\end{equation}
where $D{=}512$ matches the image cross-attention dimension. We inject degradation cues into the image branch rather than the text branch because degradation is inherently visual and instance-specific, and thus is better aligned with image-conditioned attention than with semantic text conditioning. In our default dynamic variant, the degradation token is further conditioned on the current diffusion timestep $t$ via scale-and-shift modulation:
\begin{equation}
  \mathbf{t}_{\mathrm{deg}}' = \mathrm{LN}\!\bigl(
    \mathrm{MLP}(\mathbf{d}) \odot (1 + \boldsymbol{\gamma}_t)
    + \boldsymbol{b}_t \bigr),
  \label{eq:deg_dynamic}
\end{equation}
where $[\boldsymbol{\gamma}_t,\, \boldsymbol{b}_t] = \mathrm{MLP}_t(\mathrm{PE}(t))$, $\mathrm{PE}(t)$ is a sinusoidal timestep embedding, and $\odot$ denotes element-wise multiplication. This allows the conditioning strength to adapt across the denoising trajectory. The full adapter is compact relative to the backbone, and its parameter and runtime overhead is quantified in Sec.~\ref{sec:overhead}.

Let $\mathbf{H}_{\mathrm{img}} \in \mathbb{R}^{N \times D}$ denote the image hidden states produced by the RAM image encoder projection, where $N$ is the number of image tokens. The degradation token is appended along the token dimension:
\begin{equation}
  \mathbf{H} = [\,\mathbf{H}_{\mathrm{img}} ;\; \mathbf{t}_{\mathrm{deg}}'\,]
  \in \mathbb{R}^{(N+1) \times D},
  \label{eq:concat}
\end{equation}
which serves as the key and value in the image cross-attention layers of both ControlNet and UNet. In this way, the model can jointly exploit semantic and degradation information without modifying the pretrained backbone architecture.

\subsection{Spatially Asymmetric Noise Injection}
Standard diffusion training applies spatially uniform Gaussian noise $\boldsymbol{\epsilon} \sim \mathcal{N}(\mathbf{0}, \mathbf{I})$ to all latent locations. However, edge and boundary regions contain critical geometric cues that are more sensitive to perturbation than flat areas. We introduce a training-time structure-aware noise modulation strategy.

We compute an edge-strength map from $\mathbf{x}_{\mathrm{lr}}$ using Sobel filtering:
\begin{equation}
  \mathbf{E}_{\mathrm{raw}} = \sqrt{(I_g * S_x)^2 + (I_g * S_y)^2},
\end{equation}
where $S_x, S_y$ are horizontal and vertical Sobel kernels. The map is normalized per image to $[0,1]$ and bilinearly interpolated to the latent resolution to obtain $\mathbf{E} \in \mathbb{R}^{1 \times h \times w}$. The sampled noise is then modulated as:
\begin{equation}
  \boldsymbol{\epsilon}' = \boldsymbol{\epsilon} \odot \bigl(1 - \lambda\,\mathbf{E}\bigr),
  \label{eq:spatial_noise}
\end{equation}
where $\lambda \in [0,1]$ controls the modulation strength (default $\lambda{=}0.6$), and $\mathbf{E}$ is broadcast across all $C$ latent channels. Edge-rich regions receive up to $\lambda$ fraction less noise amplitude (equivalently, a variance reduction of up to $\lambda^2 = 0.36$), while smooth regions retain near-standard perturbation. The noised latent is:
\begin{equation}
  \mathbf{z}_t = \sqrt{\bar{\alpha}_t}\,\mathbf{z}_0 + \sqrt{1-\bar{\alpha}_t}\,\boldsymbol{\epsilon}'.
  \label{eq:noisy_latent}
\end{equation}

\noindent\textbf{Train--inference consistency.}\quad
Since SANI modifies the forward-process noise, the training distribution deviates slightly from the standard isotropic Gaussian assumption underlying the DDPM reverse process. We note that the deviation is bounded: with $\lambda{=}0.6$, the per-element noise amplitude scale lies in $[0.4,\,1.0]$, i.e., the modulated noise remains in the same order of magnitude as the original. At inference time, we keep the standard reverse diffusion process and do not modify the sampling scheduler.

To analyze why this mismatch is acceptable, we note three complementary factors. First, the modulated noise distribution is still zero-mean and spatially smooth, deviating only in local variance rather than in distribution shape; this is a much milder perturbation than, e.g., replacing Gaussian noise with a fundamentally different distribution. Second, the denoising score function learned by large-scale diffusion models is known to be locally robust to small perturbations in the noise level schedule~\cite{IDDPM}, because the model is trained to denoise across a wide range of noise magnitudes. Third, the mismatch is concentrated in high-gradient (edge) regions where the noise reduction is strongest; these regions are also where the model benefits most from reduced perturbation during training, as structural cues are better preserved. Empirically, the pretrained diffusion backbone exhibits sufficient robustness to this mild distributional shift, and the benefit of preserving structural cues during training outweighs the small mismatch at inference time, as confirmed by our ablation study. 

SANI introduces no additional loss terms, auxiliary labels, or architectural changes; it only modifies the training-time noise distribution. As a result, the model is biased toward preserving structural regions without suppressing stochastic texture generation in smoother areas.

\subsection{Training and Inference}
During training, both modules are jointly enabled. For each training pair $(\mathbf{x}_{\mathrm{lr}}, \mathbf{x}_{\mathrm{hr}})$, the degradation descriptor $\mathbf{d}$ and edge-strength map $\mathbf{E}$ are extracted from $\mathbf{x}_{\mathrm{lr}}$. The degradation token is appended to the image hidden states via Eq.~\eqref{eq:concat}, and the latent noise is spatially modulated via Eq.~\eqref{eq:spatial_noise}. The training objective is:
\begin{equation}
  \mathcal{L} = \mathbb{E}_{\mathbf{z}_0,\, \boldsymbol{\epsilon}',\, t}
  \Bigl[\,
    \bigl\|\,\hat{\boldsymbol{\epsilon}}_\theta\bigl(\mathbf{z}_t,\, t,\, \mathbf{c}_{\mathrm{text}},\, \mathbf{x}_{\mathrm{lr}},\, \mathbf{H}\bigr)
    - \boldsymbol{\epsilon}'\,\bigr\|_2^2
  \,\Bigr],
  \label{eq:loss}
\end{equation}
where $\mathbf{c}_{\mathrm{text}}$ is the text embedding from the frozen text encoder and $\mathbf{x}_{\mathrm{lr}}$ is fed through ControlNet as spatial conditioning. We jointly optimize the ControlNet, the image cross-attention modules of the UNet, and the degradation token adapter, while keeping the pretrained VAE, text encoder, and RAM image tagging model frozen. Training uses AdamW ($\beta_1{=}0.9$, $\beta_2{=}0.999$, weight decay $10^{-2}$) with a constant learning rate of $5\times10^{-5}$ after 500 warmup steps. The per-GPU batch size is 2 across 6 GPUs with gradient accumulation over 8 steps, yielding an effective batch size of 96. Mixed-precision (fp16) is used throughout, and gradients are clipped to a maximum norm of 1.0.

During inference, degradation descriptor extraction and token injection are retained to ensure train--test consistency. In the dynamic variant (our default), the degradation token is re-computed at each denoising step according to the current timestep; this introduces only limited additional overhead, as the adapter contains only two linear layers (see Sec.~\ref{sec:overhead}). Since SANI is a training-time strategy only, it introduces no additional inference cost. Both modules are lightweight add-ons and can be incorporated into diffusion-based SR frameworks with only minor modifications to the conditioning pipeline; their overhead is quantified in Sec.~\ref{sec:overhead}.

\section{Experiments}
\label{sec:experiments}

\subsection{Experimental Setup}

\noindent\textbf{Datasets.}
We conduct experiments on two widely used evaluation benchmarks: DIV2K~\cite{DIV2K}, which contains 800 high-resolution images with diverse scenes and rich textures, and RealSR~\cite{RealSR}, which provides real-world LR-HR image pairs captured with different camera focal lengths. For training, we use the large-scale LSDIR dataset~\cite{LSDIR}. Following common practice in real-world SR, the corresponding LR images are synthesized using the Real-ESRGAN degradation pipeline~\cite{RealESRGAN}, which simulates realistic degradations such as blur, noise, and JPEG compression.

\noindent\textbf{Metrics.}
We evaluate restoration quality using both reference-based and no-reference metrics. Reference-based metrics include PSNR, SSIM~\cite{SSIM}, and LPIPS~\cite{LPIPS}. No-reference perceptual metrics include NIQE~\cite{NIQE}, NRQM, PI~\cite{PI}, CLIP-IQA~\cite{CLIPIQA}, and MUSIQ~\cite{MUSIQ}. Since real-world SR involves unknown and spatially non-uniform degradations, reference-based fidelity does not always align with perceptual restoration quality. We therefore report both types of metrics for a comprehensive evaluation.

\noindent\textbf{Compared Methods.}
We compare our method with recent state-of-the-art real-world SR approaches, including both regression/GAN-based and diffusion-based methods: Real-ESRGAN~\cite{RealESRGAN}, SeeSR~\cite{SeeSR}, StableSR~\cite{StableSR}, ResShift~\cite{ResShift}, PASD~\cite{PASD}, DiffBIR~\cite{DiffBIR}, OSEDiff~\cite{OSEDiff}, InvSR~\cite{InvSR} and PiSA-SR~\cite{PiSA-SR}.

\subsection{Comparison with State-of-the-Art}

We compare our method with state-of-the-art real-world SR methods on DIV2K and RealSR. Quantitative results are reported in Table~\ref{tab:main}.

\begin{table*}[t]
\centering
\caption{
Quantitative comparison on DIV2K and RealSR ($\times4$ SR).
$\uparrow$ indicates higher is better and $\downarrow$ indicates lower is better.
Best results are shown in \textbf{bold}.
}
\label{tab:main}
\setlength{\tabcolsep}{5pt}
\renewcommand{\arraystretch}{1.08}
\small
\resizebox{\textwidth}{!}{
\begin{tabular}{c|l|cccccccc}
\toprule
\textbf{Datasets} & \textbf{Methods} & \multicolumn{8}{c}{\textbf{Metrics}} \\
\cmidrule(lr){3-10}
& & PSNR$\uparrow$ & SSIM$\uparrow$ & LPIPS$\downarrow$ & NIQE$\downarrow$ & NRQM$\uparrow$ & PI$\downarrow$ & CLIP-IQA$\uparrow$ & MUSIQ$\uparrow$ \\
\midrule

\multirow{10}{*}{\textit{DIV2K}}
& Real-ESRGAN~\cite{RealESRGAN}  & 26.65 & 0.758 & 0.18 & 3.591 & 6.296 & 3.722 & 0.565 & 64.661 \\
& SeeSR~\cite{SeeSR}  & 26.457 & 0.728 & 0.172 & 3.5 & 6.54 & 3.566 & 0.679 & 68.709 \\
& StableSR~\cite{StableSR}  & 25.86 & \textbf{0.77} & 0.216 & 4.813 & 5.258 & 4.804 & 0.551 & 59.106 \\
& ResShift~\cite{ResShift}   & \textbf{27.176} & 0.767 & \textbf{0.135} & 5.127 & 6.366 & 4.371 & 0.675 & 64.95 \\
& PASD~\cite{PASD}          & 26.536 & 0.725 & 0.185 & 3.381 & 6.385 & 3.566 & 0.602 & 66.623 \\
& DiffBIR~\cite{DiffBIR}     & 21.428 & 0.562 & 0.316 & 3.054 & 6.797 & 3.138 & \textbf{0.721} & 69.445 \\
& OSEDiff~\cite{OSEDiff}      & 24.629 & 0.694 & 0.195 & 3.546 & 6.6 & 3.528 & 0.697 & 69.368 \\
& InvSR~\cite{InvSR}            & 24.148 & 0.697 & 0.198 & 3.542 & 6.661 & 3.449 & 0.692 & 69.175 \\
& PiSA-SR~\cite{PiSA-SR}         & 25.016 & 0.693 & 0.176 & 3.36 & 6.728 & 3.325 & 0.72 & \textbf{70.489} \\
\cmidrule(lr){2-10}
& Ours                              & 22.762 & 0.62 & 0.274 & \textbf{2.825} & \textbf{6.813} & \textbf{3.033} & 0.714 & 69.615 \\
\midrule

\multirow{10}{*}{\textit{RealSR}}
& Real-ESRGAN~\cite{RealESRGAN}  & 25.845 & 0.774 & 0.215 & 4.679 & 5.868 & 4.489 & 0.49 & 59.696 \\
& SeeSR~\cite{SeeSR}  & 25.799 & 0.741 & 0.223 & 4.005 & 6.469 & 3.852 & 0.675 & 66.983 \\
& StableSR~\cite{StableSR}  & 25.071 & \textbf{0.78} & \textbf{0.198} & 6.158 & 4.64 & 5.809 & 0.453 & 52.429 \\
& ResShift~\cite{ResShift}   & 25.453 & 0.725 & 0.266 & 7.349 & 6.202 & 5.55 & 0.585 & 56.189 \\
& PASD~\cite{PASD}          & \textbf{26.792} & 0.762 & 0.215 & 4.867 & 5.546 & 4.77 & 0.521 & 58.689 \\
& DiffBIR~\cite{DiffBIR}     & 19.835 & 0.514 & 0.396 & \textbf{3.709} & \textbf{6.708} & 3.553 & \textbf{0.693} & 65.01 \\
& OSEDiff~\cite{OSEDiff}      & 25.254 & 0.736 & 0.23 & 4.342 & 6.532 & 3.969 & 0.677 & 67.597 \\
& InvSR~\cite{InvSR}            & 24.928 & 0.734 & 0.211 & 4.75 & 6.672 & 4.06 & 0.671 & 66.371 \\
& PiSA-SR~\cite{PiSA-SR}         & 25.587 & 0.749 & 0.208 & 4.388 & 6.421 & 4.033 & 0.668 & 67.995 \\
\cmidrule(lr){2-10}
& Ours                              & 22.69 & 0.641 & 0.299 & \textbf{3.709} & 6.694 & \textbf{3.532} & 0.681 & \textbf{69.597} \\
\bottomrule
\end{tabular}
}
\end{table*}

\noindent\textbf{Objective Evaluation.}
As shown in Table~\ref{tab:main}, our method achieves competitive perceptual restoration performance on both benchmarks. In particular, it achieves the best or tied-best results on NIQE, NRQM, PI, and MUSIQ across the two datasets, indicating that the proposed degradation-aware conditioning and structure-preserving noise design are effective for improving perceptual quality under complex degradations. While our method is not optimized for distortion-oriented metrics such as PSNR and SSIM, it provides a favorable perception-distortion trade-off and yields visually more realistic restoration results.

\noindent\textbf{Discussion: Perception--Distortion Trade-off.}
The lower PSNR and SSIM of our method relative to regression-based baselines (e.g., ResShift, Real-ESRGAN) is consistent with the well-established perception--distortion trade-off~\cite{PerceptionDistortion}: methods that generate sharper, more realistic textures tend to deviate further from the pixel-level reference, resulting in lower distortion-oriented scores. Our method is explicitly designed for the high-perceptual-quality operating point, targeting applications such as photo enhancement, content creation, and display-oriented SR, where visual realism is prioritized over strict pixel fidelity. The improvement in no-reference metrics (NIQE, NRQM, PI, CLIP-IQA, MUSIQ) reflects more natural texture statistics and sharper structural details, which better align with human visual perception in the presence of complex real-world degradations. For scenarios where high fidelity is required, the baseline variant (without DTI and SANI) already achieves higher PSNR/SSIM, as shown in the ablation study.

\begin{figure*}[ht]
    \centering
    \includegraphics[width=1\textwidth]{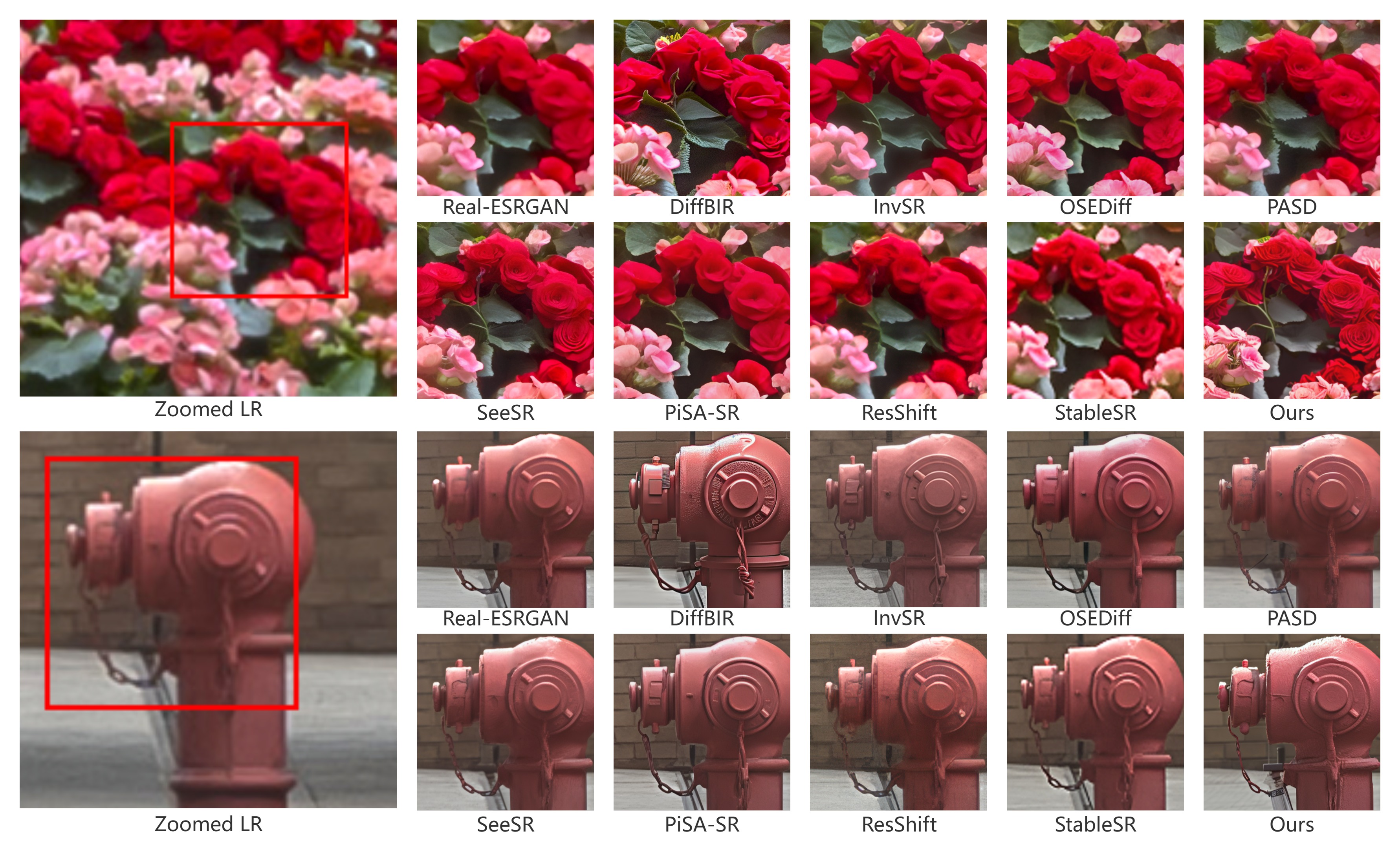}
    \caption{Visual results of different methods on two typical real-world examples from the RealSR dataset.}
    \label{fig:qualitative_2}
\end{figure*}

\noindent\textbf{Visual Comparison.}
Fig.~\ref{fig:qualitative_1} and Fig.~\ref{fig:qualitative_2} further provide qualitative comparisons on representative LR images from DIV2K and RealSR. Compared with prior methods, our approach recovers sharper edges, finer textures, and more semantically consistent details, particularly in regions affected by complex or spatially non-uniform degradations. These visual results are consistent with the quantitative findings and further support the benefit of combining degradation-aware conditioning with structure-preserving noise design in diffusion-based real-world SR.

\subsection{Complexity and Runtime Analysis}
\label{sec:overhead}

We further quantify the computational overhead of the proposed modules in terms of parameter count and inference runtime. DTI introduces only a lightweight degradation-token adapter implemented by a two-layer MLP, together with one additional conditioning token in the image cross-attention branch. By contrast, SANI is parameter-free and only modifies the noise generation process during training, introducing no extra inference-time computation. Therefore, the inference-time overhead of the full model is effectively determined by DTI.

It is worth noting that diffusion-based SR is inherently sampling-intensive, and thus the absolute inference cost is mainly dominated by the backbone itself. In this context, the lightweight nature of our design should be understood as the incremental overhead introduced over the adopted diffusion SR backbone, rather than the absolute runtime of the full diffusion pipeline. As shown in Table~\ref{tab:overhead}, DTI introduces only 0.217M additional parameters, corresponding to a 0.01\% increase over the baseline. The measured latency increases by only 0.10 s/image after introducing DTI (16.58 vs.\ 16.67 s/image), indicating very limited runtime overhead in practice.

\begin{table}[t]
\centering
\caption{Overhead analysis of the proposed modules. SANI is only used during training and introduces no extra inference cost.}
\label{tab:overhead}
\renewcommand{\arraystretch}{1.05}
\small
\resizebox{\columnwidth}{!}{
\begin{tabular}{lcccc}
\toprule
\textbf{Method} & \textbf{Params (M)} & \textbf{$\Delta$ Params} & \textbf{Time (s/img)} & \textbf{$\Delta$ Time (s)} \\
\midrule
Baseline       & 2249.91 & --      & 16.58 & -- \\
+ DTI          & 2250.13 & +0.01\% & 16.67 & +0.10 \\
+ SANI         & 2249.91 & +0.00\% & 16.58 & +0.00 \\
+ DTI + SANI   & 2250.13 & +0.01\% & 16.67 & +0.10 \\
\bottomrule
\end{tabular}
}
\end{table}
\subsection{Ablation Study}
\label{sec:ablation}

We conduct ablation experiments on both DIV2K and RealSR to analyze the individual and joint effects of Degradation-aware Token Injection (DTI) and Spatially Asymmetric Noise Injection (SANI). For fair comparison, all variants use the same backbone, training data, inference settings, and checkpoint selection protocol. Results are reported in Table~\ref{tab:ablation}.

\begin{table*}[t]
\centering
\caption{
Ablation study on the proposed components across DIV2K and RealSR datasets.
DTI: Degradation-aware Token Injection. SANI: Spatially Asymmetric Noise Injection.
$\uparrow$ indicates higher is better and $\downarrow$ indicates lower is better.
}
\label{tab:ablation}
\setlength{\tabcolsep}{8pt}
\renewcommand{\arraystretch}{1.1}
\small
\resizebox{\textwidth}{!}{
\begin{tabular}{c|cc|cccccccc}
\toprule
\textbf{Datasets} & \textbf{DTI} & \textbf{SANI} & PSNR$\uparrow$ & SSIM$\uparrow$ & LPIPS$\downarrow$ & NIQE$\downarrow$ & NRQM$\uparrow$ & PI$\downarrow$ & CLIP-IQA$\uparrow$ & MUSIQ$\uparrow$ \\
\midrule

\multirow{4}{*}{\textit{DIV2K}}
& \xmark & \xmark & \textbf{25.862} & \textbf{0.717} & \textbf{0.196} & 3.604 & 6.464 & 3.641 & 0.676 & 68.449 \\
& \checkmark & \xmark & 23.677 & 0.644 & 0.280 & 3.267 & 6.624 & 3.373 & \textbf{0.730} & 69.355 \\
& \xmark & \checkmark & 25.066 & 0.697 & 0.199 & 3.302 & 6.681 & 3.357 & 0.691 & 69.203 \\
\cmidrule(lr){2-11}
& \checkmark & \checkmark & 22.762 & 0.620 & 0.274 & \textbf{2.825} & \textbf{6.813} & \textbf{3.033} & 0.714 & \textbf{69.615} \\

\midrule

\multirow{4}{*}{\textit{RealSR}}
& \xmark & \xmark & \textbf{25.017} & \textbf{0.726} & \textbf{0.242} & 4.278 & 6.449 & 3.992 & 0.684 & 68.236 \\
& \checkmark & \xmark & 23.530 & 0.671 & 0.299 & 4.336 & 6.430 & 3.990 & 0.681 & 67.498 \\
& \xmark & \checkmark & 23.848 & 0.692 & 0.255 & 3.872 & 6.635 & 3.667 & \textbf{0.701} & 68.558 \\
\cmidrule(lr){2-11}
& \checkmark & \checkmark & 22.690 & 0.641 & 0.299 & \textbf{3.709} & \textbf{6.694} & \textbf{3.532} & 0.681 & \textbf{69.597} \\

\bottomrule
\end{tabular}
}
\end{table*}

\noindent\textbf{Analysis of the Perception-Distortion Trade-off.}
As shown in Table~\ref{tab:ablation}, we observe a typical perception-distortion trade-off across all variants. While the baseline model (\xmark, \xmark) achieves higher PSNR and SSIM, it tends to produce over-smoothed results with weaker perceptual quality. In contrast, the proposed modules improve several no-reference quality metrics, which better correlate with visual realism in real-world SR scenarios.

\noindent\textbf{Effect of DTI.}
Introducing Degradation-aware Token Injection (DTI) provides explicit degradation cues to the diffusion model. On DIV2K, DTI improves perception-oriented metrics such as CLIP-IQA and MUSIQ, indicating that explicit degradation conditioning helps the model synthesize more realistic textures under complex degradations. On RealSR, DTI alone does not yield consistent gains across all perceptual metrics (e.g., NIQE slightly increases from 4.278 to 4.336), suggesting that degradation-aware conditioning alone is insufficient without complementary structural preservation; the two modules are designed to work in synergy.

\noindent\textbf{Effect of SANI.}
Spatially Asymmetric Noise Injection (SANI) is designed to preserve local geometry during diffusion training. Comparing (\xmark, \xmark) with (\xmark, \checkmark), SANI consistently improves perceptual metrics such as NIQE, NRQM, and PI across both datasets. For example, on RealSR, SANI reduces NIQE from 4.278 to 3.872. This suggests that structure-aware noise injection helps the model recover sharper edges and more natural local details.

\noindent\textbf{Joint Effect of DTI and SANI.}
The full model (\checkmark, \checkmark) achieves the strongest overall perceptual performance. On DIV2K, it obtains the best NIQE (2.825), NRQM (6.813), PI (3.033), and MUSIQ (69.615). On RealSR, it achieves the best or tied-best results on NIQE (3.709), PI (3.532), and MUSIQ (69.597). These results indicate that DTI and SANI are complementary: DTI provides explicit degradation-aware guidance, while SANI improves local structural preservation. Their combination leads to more visually pleasing restoration results under complex real-world degradations.

\section{Conclusion}
\label{sec:conclusion}

We presented a degradation-aware and structure-preserving diffusion framework for real-world image super-resolution. By incorporating Degradation-aware Token Injection (DTI) and Spatially Asymmetric Noise Injection (SANI), the proposed method enables more explicit degradation conditioning and better structural preservation during diffusion training. Extensive experiments on DIV2K and RealSR show that our method achieves competitive perceptual restoration quality and a favorable perception-distortion trade-off against recent baselines. These results suggest that explicit degradation modeling and structure-aware diffusion design are both important for advancing real-world SR.

\begin{figure}[ht!]
\centering
\includegraphics[width=\linewidth]{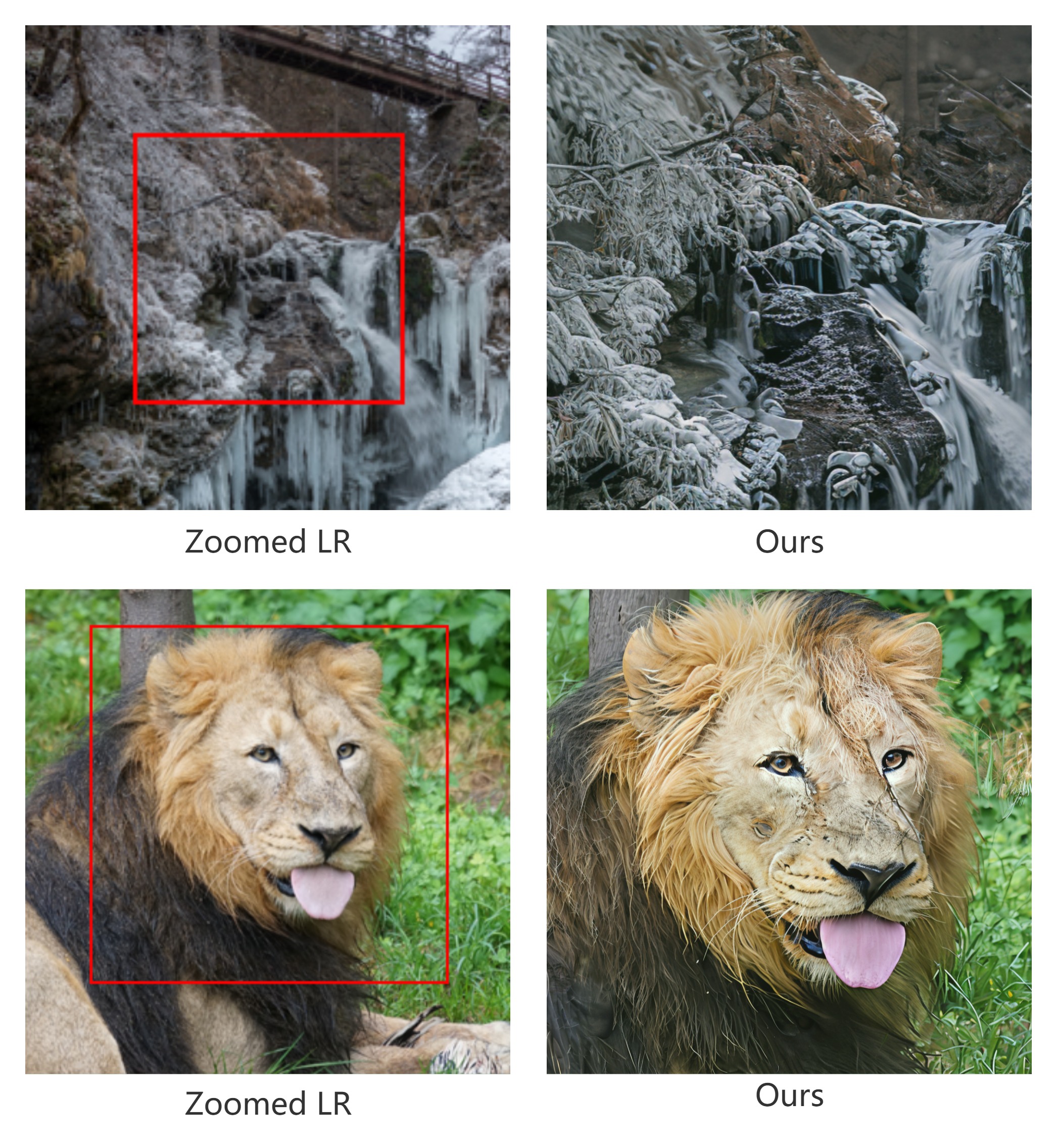}
\caption{Representative failure case of the proposed method.}
\label{fig:fail_case}
\end{figure}

\noindent\textbf{Limitations.}
Despite its effectiveness, our method inherits the relatively high inference cost of multi-step diffusion models, which restricts its use in efficiency-sensitive scenarios. In addition, the hand-crafted degradation descriptor and edge-guided noise modulation may be insufficient for highly complex, mixed, or previously unseen real-world degradations. As illustrated in Fig.~\ref{fig:fail_case}, our method can produce over-sharpened edges or hallucinated textures under such challenging inputs. A dedicated sensitivity analysis of descriptor design choices and the SANI modulation strength $\lambda$ would further strengthen the empirical evidence, and we leave this as future work. Furthermore, the current study validates the proposed modules primarily on the adopted SeeSR-based diffusion SR framework; extending the evaluation to a broader range of diffusion backbones is an important direction for future work. We will also investigate learned degradation representations and more efficient diffusion formulations to further improve robustness and inference efficiency.

{
    \small
    \bibliographystyle{ieeenat_fullname}
    \bibliography{main}
}


\end{document}